\documentclass[letterpaper, 10 pt, conference]{ieeeconf}  
\pdfoutput=1

\IEEEoverridecommandlockouts                              

\usepackage{graphics} 
\usepackage{amsmath} 
\usepackage{amssymb}  
\usepackage{graphicx}
\usepackage{hyperref}
\usepackage{subcaption}
\usepackage{graphicx}
\usepackage{mathtools}
\usepackage{dsfont}
\usepackage{float}
\usepackage{makecell}
\usepackage{authblk}

\renewcommand{\baselinestretch}{1.0}

\makeatletter
\renewcommand\AB@affilsepx{\hspace{1in} \protect\Affilfont}
\makeatother

\newcommand\blfootnote[1]{%
  \begingroup
  \renewcommand\thefootnote{}\footnote{#1}%
  \addtocounter{footnote}{-1}%
  \endgroup
}

\DeclareMathOperator{\E}{\mathbb{E}}
\newcommand\norm[1]{\left\lVert#1\right\rVert}

\title{\LARGE \bf
Overcoming Exploration in Reinforcement Learning \\ with Demonstrations
}

\author{Ashvin Nair$^{12}$, Bob McGrew$^{1}$, Marcin Andrychowicz$^{1}$, Wojciech Zaremba$^1$, Pieter Abbeel$^{12}$}

\begin{document}
\maketitle
\blfootnote{$\;^1$ OpenAI, $^2$ University of California, Berkeley.}

\thispagestyle{empty}
\pagestyle{empty}

\vspace{-10pt}
\begin{abstract}
Exploration in environments with sparse rewards has been a persistent problem in reinforcement learning (RL). Many tasks are natural to specify with a sparse reward, and manually shaping a reward function can result in suboptimal performance. However, finding a non-zero reward is exponentially more difficult with increasing task horizon or action dimensionality. This puts many real-world tasks out of practical reach of RL methods. In this work, we use demonstrations to overcome the exploration problem and successfully learn to perform long-horizon, multi-step robotics tasks with continuous control such as stacking blocks with a robot arm. Our method, which builds on top of Deep Deterministic Policy Gradients and Hindsight Experience Replay, provides an order of magnitude of speedup over RL on simulated robotics tasks. It is simple to implement and makes only the additional assumption that we can collect a small set of demonstrations. Furthermore, our method is able to solve tasks not solvable by either RL or behavior cloning alone, and often ends up outperforming the demonstrator policy.

\end{abstract}

\section{Introduction}

RL has found significant success in decision making for solving games, so what makes it more challenging to apply in robotics? A key difference is the difficulty of exploration, which comes from  the choice of reward function and complicated environment dynamics. In games, the reward function is usually given and can be directly optimized. In robotics, we often desire behavior to achieve some binary objective (e.g., move an object to a desired location or achieve a certain state of the system) which naturally induces a sparse reward. Sparse reward functions are easier to specify and recent work suggests that learning with a sparse reward results in learned policies that perform the desired objective instead of getting stuck in local optima \cite{andrychowicz2017her, vecerik17ddpgfd}. However, exploration in an environment with sparse reward is difficult since with random exploration, the agent rarely sees a reward signal.

The difficulty posed by a sparse reward is exacerbated by the complicated environment dynamics in robotics. For example, system dynamics around contacts are difficult to model and induce a sensitivity in the system to small errors. Many robotics tasks also require executing multiple steps successfully over a long horizon, involve high dimensional control, and require generalization to varying task instances. These conditions further result in a situation where the agent so rarely sees a reward initially that it is not able to learn at all.

All of the above means that random exploration is not a tenable solution. Instead, in this work we show that we can use demonstrations as a guide for our exploration. To test our method, we solve the problem of stacking several blocks at a given location from a random initial state. Stacking blocks has been studied before in the literature \cite{deisenroth2011blocks, duan2017oneshotimitation} and exhibits many of the difficulties mentioned: long horizons, contacts, and requires generalizing to each instance of the task. We limit ourselves to 100 human demonstrations collected via teleoperation in virtual reality. Using these demonstrations, we are able to solve a complex robotics task in simulation that is beyond the capability of both reinforcement learning and imitation learning.

The primary contribution of this paper is to show that demonstrations can be used with reinforcement learning to solve complex tasks where exploration is difficult. We introduce a simple auxiliary objective on demonstrations, a method of annealing away the effect of the demonstrations when the learned policy is better than the demonstrations, and a method of resetting from demonstration states that significantly improves and speeds up training policies.
By effectively incorporating demonstrations into RL, we short-circuit the random exploration phase of RL and reach nonzero rewards and a reasonable policy early on in training. Finally, we extensively evaluate our method against other commonly used methods, such as initialization with learning from demonstrations and fine-tuning with RL, and show that our method significantly outperforms them.

\begin{figure*}[t]
    \vspace{6pt}
    \centering
    \includegraphics[width=1.0\linewidth]{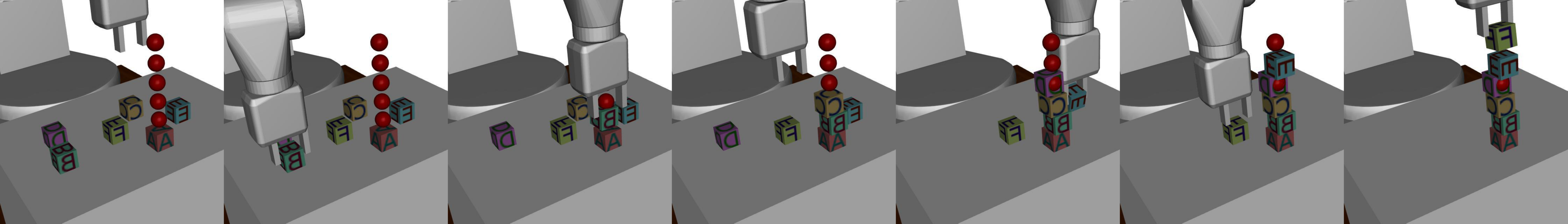}
    
    \vspace{0.1cm}
    
    \includegraphics[width=1.0\linewidth]{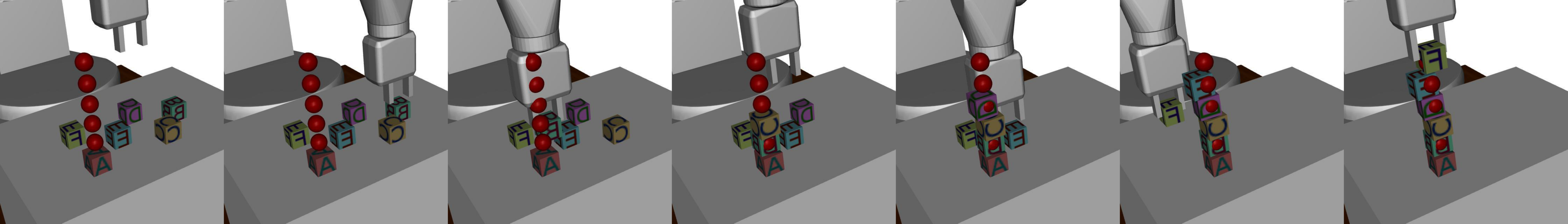}
    \caption{We present a method using reinforcement learning to solve the task of block stacking shown above. The robot starts with 6 blocks labelled A through F on a table in random positions and a target position for each block. The task is to move each block to its target position. The targets are marked in the above visualization with red spheres which do not interact with the environment. These targets are placed in order on top of block A so that the robot forms a tower of blocks. This is a complex, multi-step task where the agent needs to learn to successfully manage multiple contacts to succeed. Frames from rollouts of the learned policy are shown. A video of our experiments can be found at:  \url{http://ashvin.me/demoddpg-website} }%
    \vspace{-3pt}
    \label{fig:fig1}
\end{figure*}

\section{Related Work}

Learning methods for decision making problems such as robotics largely divide into two classes:  imitation learning and reinforcement learning (RL).  In imitation learning (also called learning from demonstrations) the agent receives behavior examples from an expert and attempts to solve a task by copying the expert's behavior. In RL, an agent attempts to maximize expected reward through interaction with the environment. Our work combines aspects of both to solve complex tasks.

\textbf{Imitation Learning:} Perhaps the most common form of imitation learning is behavior cloning (BC), which learns a policy through supervised learning on demonstration state-action pairs.  BC has seen success in autonomous driving \cite{pomerleau1989alvinn, bojarski2016nvidia}, quadcopter navigation \cite{giusti15trails}, locomotion \cite{nakanishi2004bipedlfd, kalakrishnan09terraintemplates}.
BC struggles outside the manifold of demonstration data. Dataset Aggregation (DA\scriptsize{GGER}\normalsize) augments the dataset by interleaving the learned and expert policy to address this problem of accumulating errors \cite{ross2011dagger}. However, DA\scriptsize{GGER}\normalsize \, is difficult to use in practice as it requires access to an expert during all of training, instead of just a set of demonstrations.

Fundamentally, BC approaches are limited because they do not take into account the task or environment. Inverse reinforcement learning (IRL) \cite{ng2000irl} is another form of imitation learning where a reward function is inferred from the demonstrations. Among other tasks, IRL has been applied to navigation \cite{ziebart2008maxent}, autonomous helicopter flight \cite{abbeel2004apprenticeship}, and manipulation \cite{finn16guidedcostlearning}. Since our work assumes knowledge of a reward function, we omit comparisons to IRL approaches.

\textbf{Reinforcement Learning:} Reinforcement learning methods have been harder to apply in robotics, but are heavily investigated because of the autonomy they could enable. Through RL, robots have learned to play table tennis \cite{peters2010reps}, swing up a cartpole, and balance a unicycle \cite{deisenroth2011pilco}. A renewal of interest in RL cascaded from success in games \cite{mnih2015human, Silver2016}, especially because of the ability of RL with large function approximators (ie. deep RL) to learn control from raw pixels. Robotics has been more challenging in general but there has been significant progress. Deep RL has been applied to manipulation tasks \cite{LevineFDA15}, grasping \cite{pinto2015supersizing, levine2016learning}, opening a door \cite{Gu2016b}, and locomotion \cite{lillicrap2015continuous, mnih2016asynchronous, schulman2015trpo}.  However, results have been attained predominantly in simulation per high sample complexity, typically caused by exploration challenges.

\textbf{Robotic Block Stacking:} Block stacking has been studied from the early days of AI and robotics as a task that encapsulates many difficulties of more complicated tasks we want to solve, including multi-step planning and complex contacts. SHRDLU \cite{winograd72shrdlr} was one of the pioneering works, but studied block arrangements only in terms of logic and natural language understanding. More recent work on task and motion planning considers both logical and physical aspects of the task \cite{Kaelbling2011, Kavraki1996, srivastava14tamp}, but requires domain-specific engineering. In this work we study how an agent can learn this task without the need of domain-specific engineering.

One RL method, PILCO \cite{deisenroth2011pilco} has been applied to a simple version of stacking blocks where the task is to place a block on a tower \cite{deisenroth2011blocks}. Methods such as PILCO based on learning forward models naturally have trouble modelling the sharply discontinuous dynamics of contacts; although they can learn to place a block, it is a much harder problem to grasp the block in the first place.  One-shot Imitation \cite{duan2017oneshotimitation} learns to stack blocks in a way that generalizes to new target configurations, but uses more than 100,000 demonstrations to train the system. A heavily shaped reward can be used to learn to stack a Lego block on another with RL \cite{popov17stacking}. In contrast, our method can succeed from fully sparse rewards and handle stacking several blocks.

\textbf{Combining RL and Imitation Learning:} 
Previous work has combined reinforcement learning with demonstrations. Demonstrations have been used to accelerate learning on classical tasks such as cart-pole swing-up and balance \cite{schaal97lfd}. This work initialized policies and (in model-based methods) initialized forward models with demonstrations. Initializing policies from demonstrations for RL has been used for learning to hit a baseball \cite{peters2008baseball} and for underactuated swing-up \cite{kober2008mp}. Beyond initialization, we show how to extract more knowledge from demonstrations by using them effectively throughout the entire training process.

Our method is closest to two recent approaches --- Deep Q-Learning From Demonstrations (DQfD) \cite{hester17dqfd} and DDPG From Demonstrations (DDPGfD) \cite{vecerik17ddpgfd} which combine demonstrations with reinforcement learning. DQfD improves learning speed on Atari, including a margin loss which encourages the expert actions to have higher Q-values than all other actions. This loss can make improving upon the demonstrator policy impossible which is not the case for our method. Prior work has previously explored improving beyond the demonstrator policy in simple environments by introducing slack variables \cite{kim2013apid}, but our method uses a learned value to actively inform the improvement. DDPGfD solves simple robotics tasks akin to peg insertion using DDPG with demonstrations in the replay buffer. In contrast to this prior work, the tasks we consider exhibit additional difficulties that are of key interest in robotics: multi-step behaviours, and generalization to varying goal states.
While previous work focuses on speeding up already solvable tasks, we show that we can extend the state of the art in RL with demonstrations by introducing new methods to incorporate demonstrations.

\section{Background}

\subsection{Reinforcement Learning}

We consider the standard Markov Decision Process framework for picking optimal actions to maximize rewards over discrete timesteps in an environment $E$. We assume that the environment is fully observable. At every timestep $t$, an agent is in a state $x_t$, takes an action $a_t$, receives a reward $r_t$, and $E$ evolves to state $x_{t+1}$. In reinforcement learning, the agent must learn a policy $a_t = \pi(x_t)$ to maximize expected returns.  We denote the return by $R_t = \sum_{i=t}^T \gamma^{(i - t)} r_i$ where $T$ is the horizon that the agent optimizes over and $\gamma$ is a discount factor for future rewards. The agent's objective is to maximize expected return from the start distribution $J = \E_{r_i, s_i \sim E, a_i \sim \pi}[R_0]$. 

A variety of reinforcement learning algorithms have been developed to solve this problem. Many involve constructing an estimate of the expected return from a given state after taking an action:
\begin{align}
    Q^\pi & (s_t, a_t) = \E_{r_i, s_i \sim E, a_i \sim \pi}[R_t|s_t, a_t] \label{eqn:Q} \\
                   & = \E_{r_t, s_{t+1} \sim E}[r_t + \gamma \E_{a_{t+1} \sim \pi}[Q^\pi(s_{t+1}, a_{t+1})]] \label{eqn:bellman}
\end{align}

\noindent We call $Q^\pi$ the action-value function. Equation \ref{eqn:bellman} is a recursive version of equation \ref{eqn:Q}, and is known as the Bellman equation. The Bellman equation allows for methods to estimate $Q$ that resemble dynamic programming.

\subsection{DDPG}

Our method combines demonstrations with one such method: Deep Deterministic Policy Gradients (DDPG) \cite{lillicrap2015continuous}. 
DDPG is an off-policy model-free reinforcement learning algorithm for continuous control which can utilize large function approximators such as neural networks. DDPG is an actor-critic method, which bridges the gap between policy gradient methods and value approximation methods for RL. At a high level, DDPG learns an action-value function (critic) by minimizing the Bellman error, while simultaneously learning a policy (actor) by directly maximizing the estimated action-value function with respect to the parameters of the policy.

Concretely, DDPG maintains an actor function $\pi(s)$ with parameters $\theta_\pi$, a critic function $Q(s, a)$ with parameters $\theta_Q$, and a replay buffer $R$ as a set of tuples $(s_t, a_t, r_t, s_{t+1})$ for each transition experienced. DDPG alternates between running the policy to collect experience and updating the parameters. Training rollouts are collected with extra noise for exploration: $a_t = \pi(s) + \mathcal{N}$, where $\mathcal{N}$ is a noise process.

During each training step, DDPG samples a minibatch consisting of $N$ tuples from $R$ to update the actor and critic networks. DDPG minimizes the following loss $L$ w.r.t. $\theta_Q$ to update the critic:
\begin{equation} \label{eq:target}
    y_i = r_i + \gamma Q(s_{i+1}, \pi(s_{i+1}))
\end{equation}
\begin{equation}
    L = \frac{1}{N}\sum_i (y_i - Q(s_i, a_i|\theta_Q))^2
\end{equation}

The actor parameters $\theta_\pi$ are updated using the policy gradient:
\begin{equation}
    \nabla_{\theta_\pi} J = \frac{1}{N}\sum_i \nabla_a Q(s, a|\theta_Q)|_{s=s_i, a=\pi(s)} \nabla_{\theta_\pi} \pi(s|\theta_\pi)|_{s_i}
\end{equation}

To stabilize learning, the $Q$ value in equation \ref{eq:target} is usually computed using a separate network (called the \emph{target} network) whose weights are an exponential average over time of the critic network.
This results in smoother target values.

Note that DDPG is a natural fit for using demonstrations. Since DDPG can be trained off-policy, we can use demonstration data as off-policy training data. We also take advantage of the action-value function $Q(s, a)$ learned by DDPG to better use demonstrations.

\subsection{Multi-Goal RL}

Instead of the standard RL setting, we train agents with parametrized goals, which lead to more general policies \cite{schaul2015uva}
and have recently been shown to make learning with sparse rewards easier \cite{andrychowicz2017her}.
Goals describe the task we expect the agent to perform in the given episode, in our case they specify the desired positions of all objects. We sample the goal $g$ at he beginning of every episode. The function approximators, here $\pi$ and $Q$, take the current goal as an additional input.

\subsection{Hindsight Experience Replay (HER)}

To handle varying task instances and parametrized goals, we use Hindsight Experience Replay (HER) \cite{andrychowicz2017her}. The key insight of HER is that even in failed rollouts where no reward was obtained, the agent can transform them into successful ones by assuming that a state it saw in the rollout was the actual goal. HER can be used with any off-policy RL algorithm assuming that for every state we can find a goal corresponding to this state (i.e. a goal which leads to a positive reward in this state).

For every episode the agent experiences, we store it in the replay buffer twice: once with the original goal pursued in the episode and once with the goal corresponding to the final state achieved in the episode, as if the agent intended on reaching this state from the very beginning.

\section{Method} \label{sec:method}

Our method combines DDPG and demonstrations in several ways to maximally use demonstrations to improve learning. We describe our method below and evaluate these ideas in our experiments.

\subsection{Demonstration Buffer} 

First, we maintain a second replay buffer $R_D$ where we store our demonstration data in the same format as $R$. In each minibatch, we draw an extra $N_D$ examples from $R_D$ to use as off-policy replay data for the update step. These examples are included in both the actor and critic update. This idea has been introduced in \cite{vecerik17ddpgfd}.

\subsection{Behavior Cloning Loss}

Second, we introduce a new loss computed only on the demonstration examples for training the actor.
\begin{equation}\label{eq:BC}
    L_{BC} = \sum_{i=1}^{N_D} \norm{\pi(s_i|\theta_\pi) - a_i}^2
\end{equation}

\noindent This loss is a standard loss in imitation learning, but we show that using it as an auxiliary loss for RL improves learning significantly. The gradient applied to the actor parameters $\theta_\pi$ is:
\begin{equation} \label{eqn:aux}
    \lambda_1 \nabla_{\theta_\pi} J - \lambda_2 \nabla_{\theta_\pi} L_{BC}
\end{equation}

\noindent (Note that we maximize $J$ and minimize $L_{BC}$.) Using this loss directly prevents the learned policy from improving significantly beyond the demonstration policy, as the actor is always tied back to the demonstrations. Next, we show how to account for suboptimal demonstrations using the learned action-value function.

\subsection{Q-Filter}\label{sec:ours}

We account for the possibility that demonstrations can be suboptimal by applying the behavior cloning loss only to states where the critic $Q(s, a)$ determines that the demonstrator action is better than the actor action:
\begin{equation}\label{eq:filter}
    L_{BC} = \sum_{i=1}^{N_D} \norm{\pi(s_i|\theta_\pi) - a_i}^2 \, \mathds{1}_{Q(s_i, a_i) > Q(s_i, \pi(s_i))}
\end{equation}

\noindent The gradient applied to the actor parameters is as in equation \ref{eqn:aux}. We label this method using the behavior cloning loss and Q-filter ``Ours'' in the following experiments.

\subsection{Resets to demonstration states}\label{sec:reset}

To overcome the problem of sparse rewards in very long horizon tasks, we reset some training episodes using states and goals from demonstration episodes. Restarts from within demonstrations expose the agent to higher reward states during training. This method makes the additional assumption that we can restart episodes from a given state, as is true in simulation.

To reset to a demonstration state, we first sample a demonstration $D = (x_0, u_0, x_1, u_1, ... x_N, u_N)$ from the set of demonstrations. We then uniformly sample a state $x_i$ from $D$.
As in HER, we use the final state achieved in the demonstration as the goal. We roll out the trajectory with the given initial state and goal for the usual number of timesteps. At evaluation time, we do not use this procedure.

We label our method with the behavior cloning loss, Q-filter, and resets from demonstration states as ``Ours, Resets'' in the following experiments.

\section{Experimental Setup}

\begin{figure*}[t]%
    \centering
    \vspace{6pt}
    \begin{subfigure}{0.3\linewidth}
        \centering
        \includegraphics[width=0.8\linewidth]{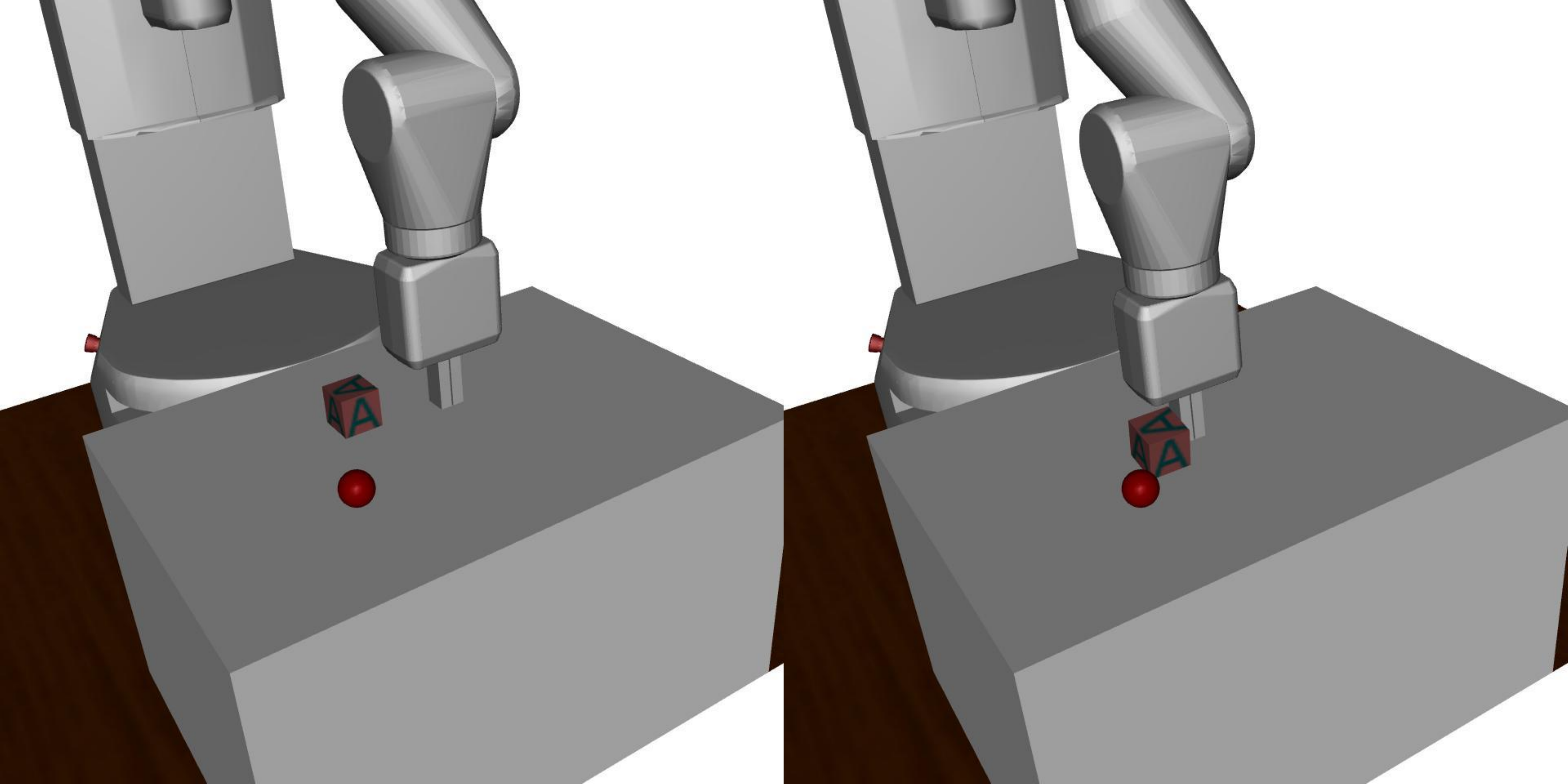}
        \includegraphics[width=1.0\linewidth]{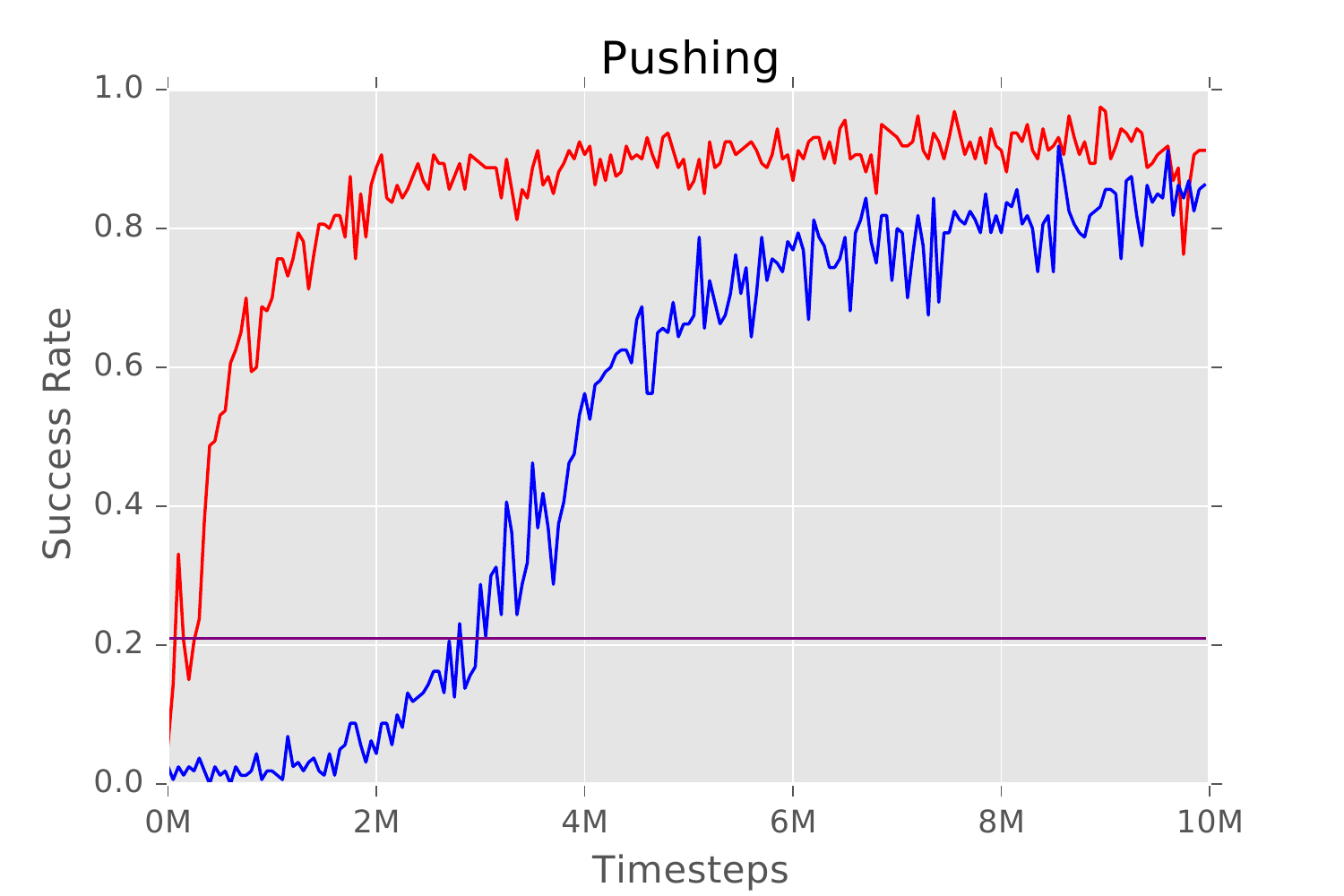}
    \end{subfigure}
    \begin{subfigure}{0.3\linewidth}
        \centering
        \includegraphics[width=0.8\linewidth]{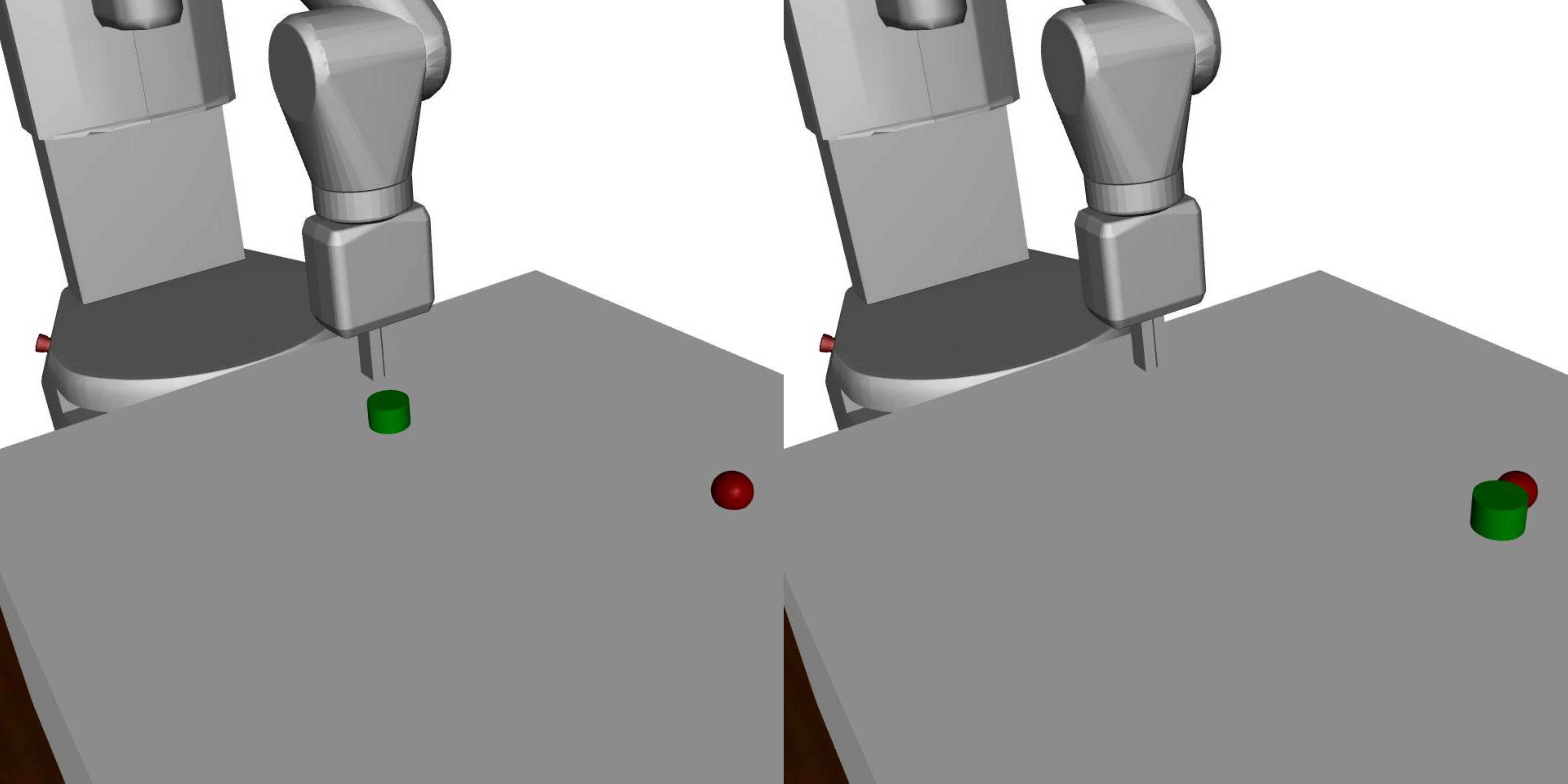}
        \includegraphics[width=1.0\linewidth]{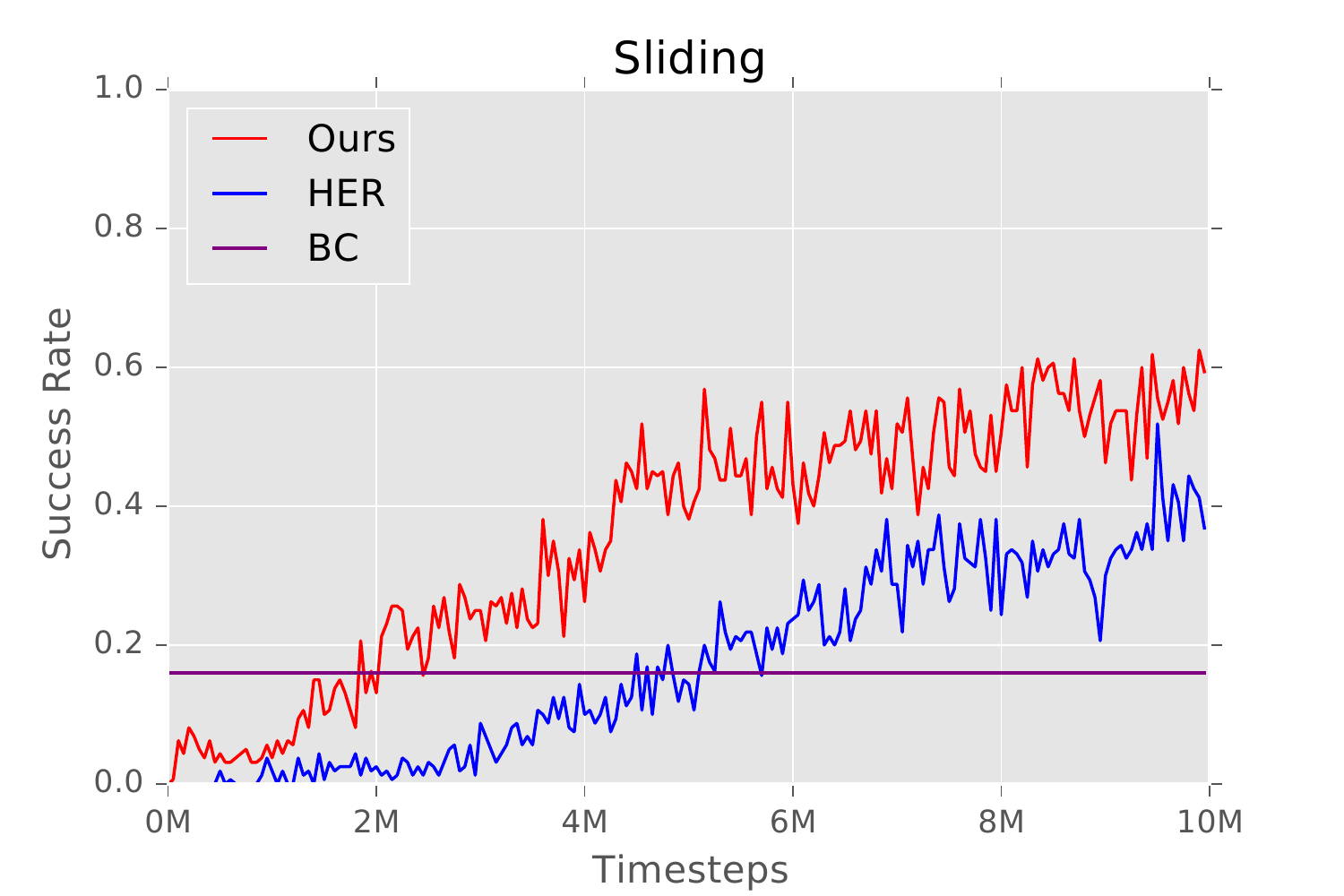}
    \end{subfigure}
    \begin{subfigure}{0.3\linewidth}
        \centering
        \includegraphics[width=0.8\linewidth]{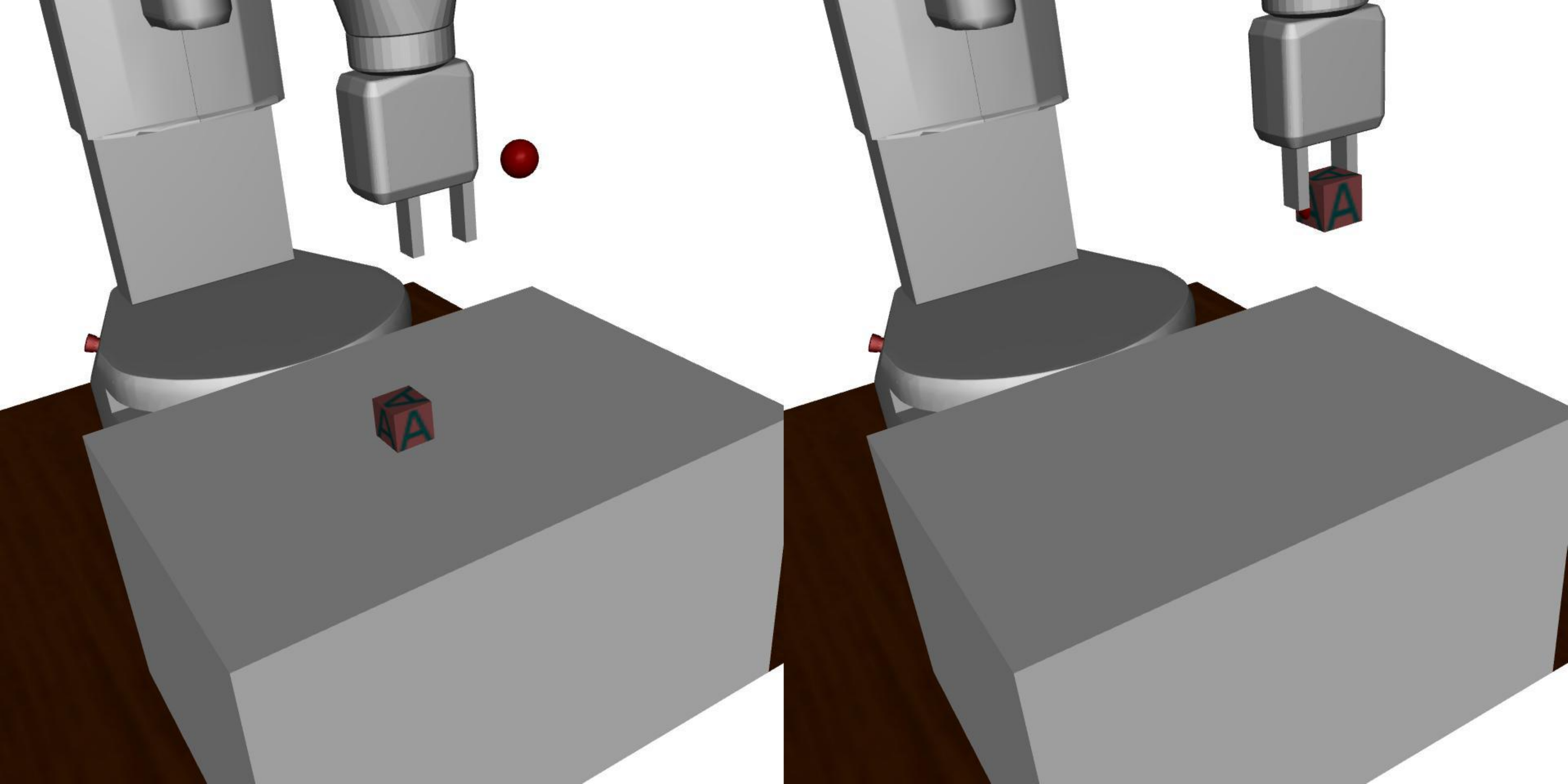}
        \includegraphics[width=1.0\linewidth]{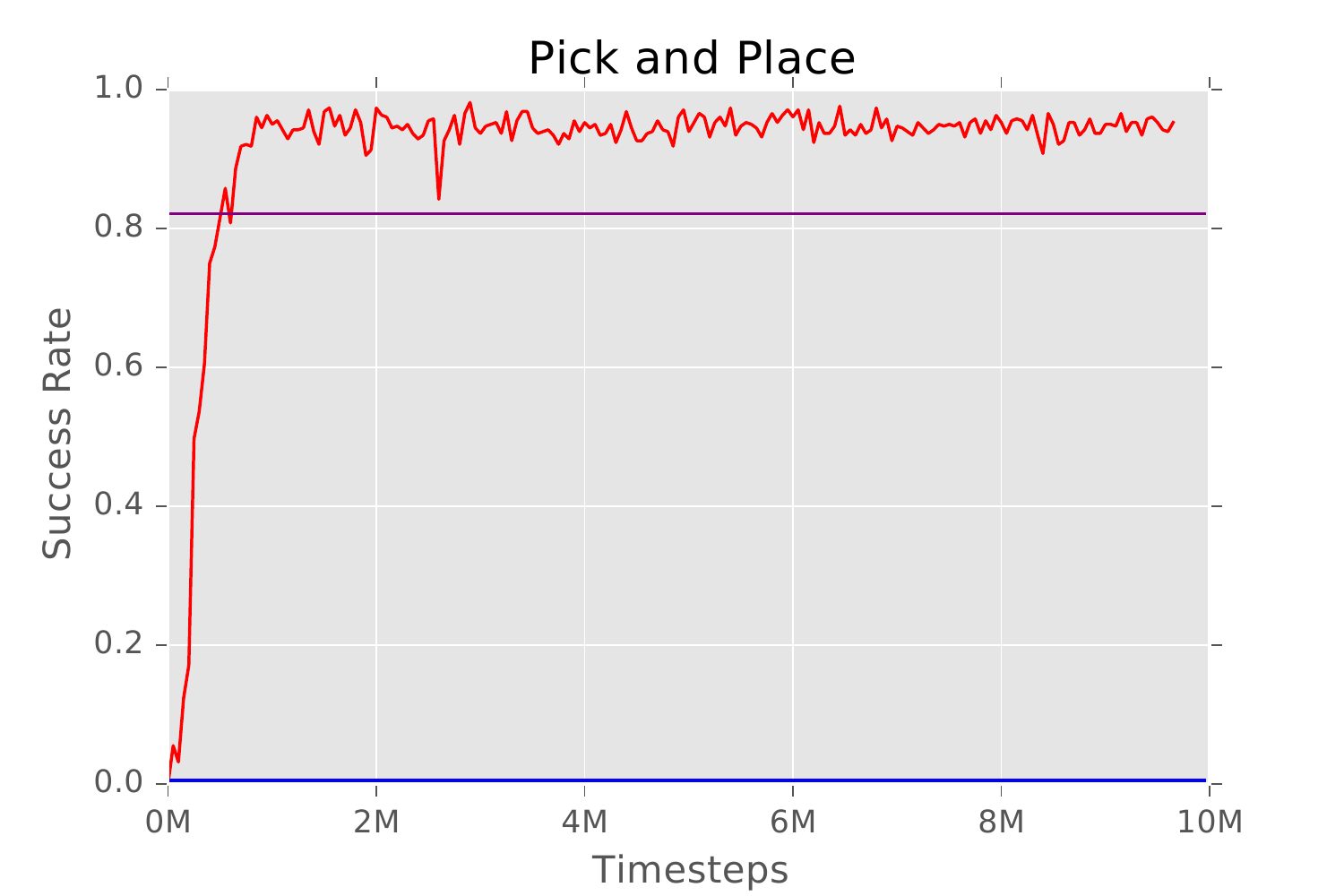}
    \end{subfigure}
    
    \caption{Baseline comparisons on tasks from \cite{andrychowicz2017her}. Frames from the learned policy are shown above each task. Our method significantly outperforms the baselines.
    On the right plot, the HER baseline always fails.}%
    \label{fig:baseline1}%
\end{figure*}

\subsection{Environments}

We evaluate our method on several simulated MuJoCo \cite{todorov12mujoco} environments. In all experiments, we use a simulated 7-DOF Fetch Robotics arm with parallel grippers to manipulate one or more objects placed on a table in front of the robot.

The agent receives the positions of the relevant objects on the table as its observations. The control for the agent is continuous and 4-dimensional: 3 dimensions that specify the desired end-effector position\footnote{In the 10cm x 10cm x 10cm cube around the current gripper position} and 1 dimension that specifies the desired distance between the robot fingers.
The agent is controlled at 50Hz frequency.

We collect demonstrations in a virtual reality environment. The demonstrator sees a rendering of the same observations as the agent, and records actions through a HTC Vive interface at the same frequency as the agent. We have the option to accept or reject a demonstration; we only accept demonstrations we judge to be mostly correct. The demonstrations are not optimal. The most extreme example is the ``sliding'' task, where only 7 of the 100 demonstrations are successful, but the agent still sees rewards for these demonstrations with HER.

\subsection{Training Details}

To train our models, we use Adam \cite{kingma2014adam} as the optimizer with learning rate $10^{-3}$. We use $N = 1024, N_D = 128, \lambda_1 = 10^{-3}, \lambda_2 = 1.0/N_D$. The discount factor $\gamma$ is 0.98. We use 100 demonstrations to initialize $R_D$. The function approximators $\pi$ and $Q$ are deep neural networks with ReLU activations and L2 regularization with the coefficient $5 \times 10^{-3}$. The final activation function for $\pi$ is tanh, and the output value is scaled to the range of each action dimension. To explore during training, we sample random actions uniformly within the action space with probability 0.1 at every step, and the noise process $\mathcal{N}$ is uniform over $\pm 10\%$ of the maximum value of each action dimension. Task-specific information, including network architectures, are provided in the next section.

\subsection{Overview of Experiments}

We perform three sets of experiments. In Sec. \ref{sec:comparison}, we provide a comparison to previous work. In Sec. \ref{sec:multistep} we solve block stacking, a difficult multi-step task with complex contacts that the baselines struggle to solve. In Sec. \ref{sec:ablations} we do ablations of our own method to show the effect of individual components.

\section{Comparison With Prior Work} \label{sec:comparison}

\subsection{Tasks}

We first show the results of our method on the simulated tasks presented in the Hindsight Experience Replay paper \cite{andrychowicz2017her}. We apply our method to three tasks:
\begin{enumerate}
    \item \textit{Pushing}. A block placed randomly on the table must be moved to a target location on the table by the robot (fingers are blocked to avoid grasping).
    \item \textit{Sliding}. A puck placed randomly on the table must be moved to a given target location. The target is outside the robot's reach so it must apply enough force that the puck reaches the target and stops due to friction.
    \item \textit{Pick-and-place}. A block placed randomly on the table must be moved to a target location in the air. Note that the original paper used a form of initializing from favorable states to solve this task. We omit this for our experiment but discuss and evaluate the initialization idea in an ablation.
\end{enumerate}

As in the prior work, we use a fully sparse reward for this task. The agent is penalized if the object is not at its goal position:
\begin{equation}
    r_t = \begin{cases}
    0, & \text{if } ||x_i - g_i|| < \delta\\
    -1, & \text{otherwise}
\end{cases}
\end{equation}

\noindent where the threshold $\delta$ is 5cm.

\subsection{Results}

Fig. \ref{fig:baseline1} compares our method to HER without demonstrations and behavior cloning. Our method is significantly faster at learning these tasks than HER, and achieves significantly better policies than behavior cloning does. Measuring the number of timesteps to get to convergence, we exhibit a 4x speedup over HER in pushing, a 2x speedup over HER in sliding, and our method solves the pick-and-place task while HER baseline cannot solve it at all.

The pick-and-place task showcases the shortcoming of RL in sparse reward settings, even with HER. In pick-and-place, the key action is to grasp the block. If the robot could manage to grasp it a small fraction of the time, HER discovers how to achieve goals in the air and reinforces the grasping behavior. However, grasping the block with random actions is extremely unlikely. Our method pushes the policy towards demonstration actions, which are more likely to succeed.

In the HER paper, HER solves the pick-and-place task by initializing half of the rollouts with the gripper grasping the block. With this addition, pick-and-place becomes the easiest of the three tasks tested. This initialization is similar in spirit to our initialization idea, but takes advantage of the fact that pick-and-place with any goal can be solved starting from a block grasped at a certain location. This is not always true (for example, if there are multiple objects to be moved) and finding such a keyframe for other tasks would be difficult, requiring some engineering and sacrificing autonomy. Instead, our method guides the exploration towards grasping the block through demonstrations. Providing demonstrations does not require expert knowledge of the learning system, which makes it a more compelling way to provide prior information.

\section{Multi-Step Experiments} \label{sec:multistep}

\subsection{Block Stacking Task}

To show that our method can solve more complex tasks with longer horizon and sparser reward, we study 
the task of block stacking in a simulated  environment as shown in Fig. \ref{fig:fig1} with the same physical properties as the previous experiments. Our experiments show that our approach can solve the task in full and learn a policy to stack 6 blocks with demonstrations and RL.  To measure and communicate various properties of our method, we also show experiments on stacking fewer blocks, a subset of the full task.

We initialize the task with blocks at 6 random locations $x_1 ... x_6$. We also provide 6 goal locations $g_1 ... g_6$. To form a tower of blocks, we let $g_1 = x_1$ and $g_i = g_{i-1} + (0, 0, 5cm)$ for $i \in 2, 3, 4, 5$.

By stacking $N$ blocks, we mean $N$ blocks reach their target locations. Since the target locations are always on top of $x_1$, we start with the first block already in position. So stacking $N$ blocks involves $N-1$ pick-and-place actions. To solve stacking $N$, we allow the agent $50 * (N-1)$ timesteps. This means that to stack 6 blocks, the robot executes 250 actions or 5 seconds.

We recorded 100 demonstrations to stack 6 blocks, and use subsets of these demonstrations as demonstrations for stacking fewer blocks. The demonstrations are not perfect; they include occasionally dropping blocks, but our method can handle suboptimal demonstrations. We still rejected more than half the demonstrations and excluded them from the demonstration data because we knocked down the tower of blocks when releasing a block.

\subsection{Rewards}

Two different reward functions are used. To test the performance of our method under fully sparse reward, we reward the agent only if all blocks are at their goal positions:
\begin{equation}
    r_t = \min_i \, \mathds{1}_{||x_i - g_i|| < \delta}
\end{equation}

\noindent The threshold $\delta$ is the size of a block, 5cm. Throughout the paper we call this the ``sparse'' reward.

To enable solving the longer horizon tasks of stacking 4 or more blocks, we use the ``step'' reward :
\begin{equation}
    r_t = -1 + \sum_i \, \mathds{1}_{||x_i - g_i|| < \delta}
\end{equation}

\noindent Note the step reward is still very sparse; the robot only sees the reward change when it moves a block into its target location. We subtract 1 only to make the reward more interpretable, as in the initial state the first block is already at its target.

Regardless of the reward type, an episode is considered successful for computing success rate if all blocks are at their goal position in their final state.

\subsection{Network architectures}

We use a 4 layer networks with 256 hidden units per layer for $\pi$ and $Q$ for the HER tasks and stacking 3 or fewer blocks. For stacking 4 blocks or more, we use an attention mechanism \cite{bahdanau14attention} for the actor and a larger network. The attention mechanism uses a 3 layer network with 128 hidden units per layer to query the states and goals with one shared head. Once a state and goal is extracted, we use a 5 layer network with 256 hidden units per layer after the attention mechanism. Attention speeds up training slightly but does not change training outcomes.

\begin{figure}[t]
    \vspace{6pt}
    \centering
    \begin{tabular}{ | c || c | c | c | c | c | }
        \hline
        Task & Ours & \makecell{Ours, \\ Resets} & BC & HER & \makecell{BC+ \\ HER}  \\ \hline
        Stack 2, Sparse & 99\% & 97\% & 65\% & 0\% & 65\%  \\ \hline
        Stack 3, Sparse & 99\% & 89\% & 1\% & 0\% & 1\%  \\ \hline
        Stack 4, Sparse & 1\% & 54\% & - & - & -  \\ \hline
        Stack 4, Step & 91\% & 73\% & 0\% & 0\% & 0\%  \\ \hline
        Stack 5, Step & 49\% & 50\% & - & - & -  \\ \hline
        Stack 6, Step & 4\% & 32\% & - & - & -  \\ \hline
    \end{tabular}
    
    \caption{Comparison of our method against baselines. The value reported is the median of the best performance (success rate) of all randomly seeded runs of each method.}
    \label{fig:scaling}%
\end{figure}

\subsection{Baselines}

We include the following methods to compare our method to baselines on stacking 2 to 6 blocks. \footnote{Because of computational constraints, we were limited to 5 random seeds per method for stacking 3 blocks, 2 random seeds per method for stacking 4 and 5 blocks, and 1 random seed per method for stacking 6 blocks. Although we are careful to draw conclusions from few random seeds, the results are consistent with our collective experience training these models. We report the median of the random seeds everywhere applicable.}

\noindent \textbf{Ours:} Refers to our method as described in section \ref{sec:ours}.

\noindent \textbf{Ours, Resets:} Refers to our method as described in section \ref{sec:ours} with resets from demonstration states (Sec.~\ref{sec:reset}).

\noindent \textbf{BC:} This method uses behavior cloning to learn a policy. Given the set of demonstration transitions $R_D$, we train the policy $\pi$ by supervised learning. Behavior cloning requires much less computation than RL. For a fairer comparison, 
we performed a large hyperparameter
sweep over various networks sizes, attention hyperparameters, and learning rates
and report the success rate achieved by the best policy found.

\noindent \textbf{HER:} This method is exactly the one described in Hindsight Experience Replay \cite{andrychowicz2017her}, using HER and DDPG.

\noindent \textbf{BC+HER:} This method first initializes a policy (actor) with BC, then finetunes the policy with RL as described above.

\subsection{Results}

We are able to learn much longer horizon tasks than the other methods, as shown in Fig. \ref{fig:scaling}. The stacking task is extremely difficult using HER without demonstrations because the chance of grasping an object using random actions is close to 0. Initializing a policy with demonstrations and then running RL also fails since the actor updates depend on a reasonable critic and although the actor is pretrained, the critic is not. The pretrained actor weights are therefore destroyed in the very first epoch, and the result is no better than BC alone. We attempted variants of this method where initially the critic was trained from replay data. However, this also fails without seeing on-policy data.

The results with sparse rewards are very encouraging. We are able to stack 3 blocks with a fully sparse reward without resetting to the states from demonstrations, and 4 blocks with a fully sparse reward if we use resetting. With resets from demonstration states and the step reward, we are able to learn a policy to stack 6 blocks. 

\begin{figure}[t]%
    \vspace{6pt}
    \centering
    \includegraphics[width=0.6\linewidth]{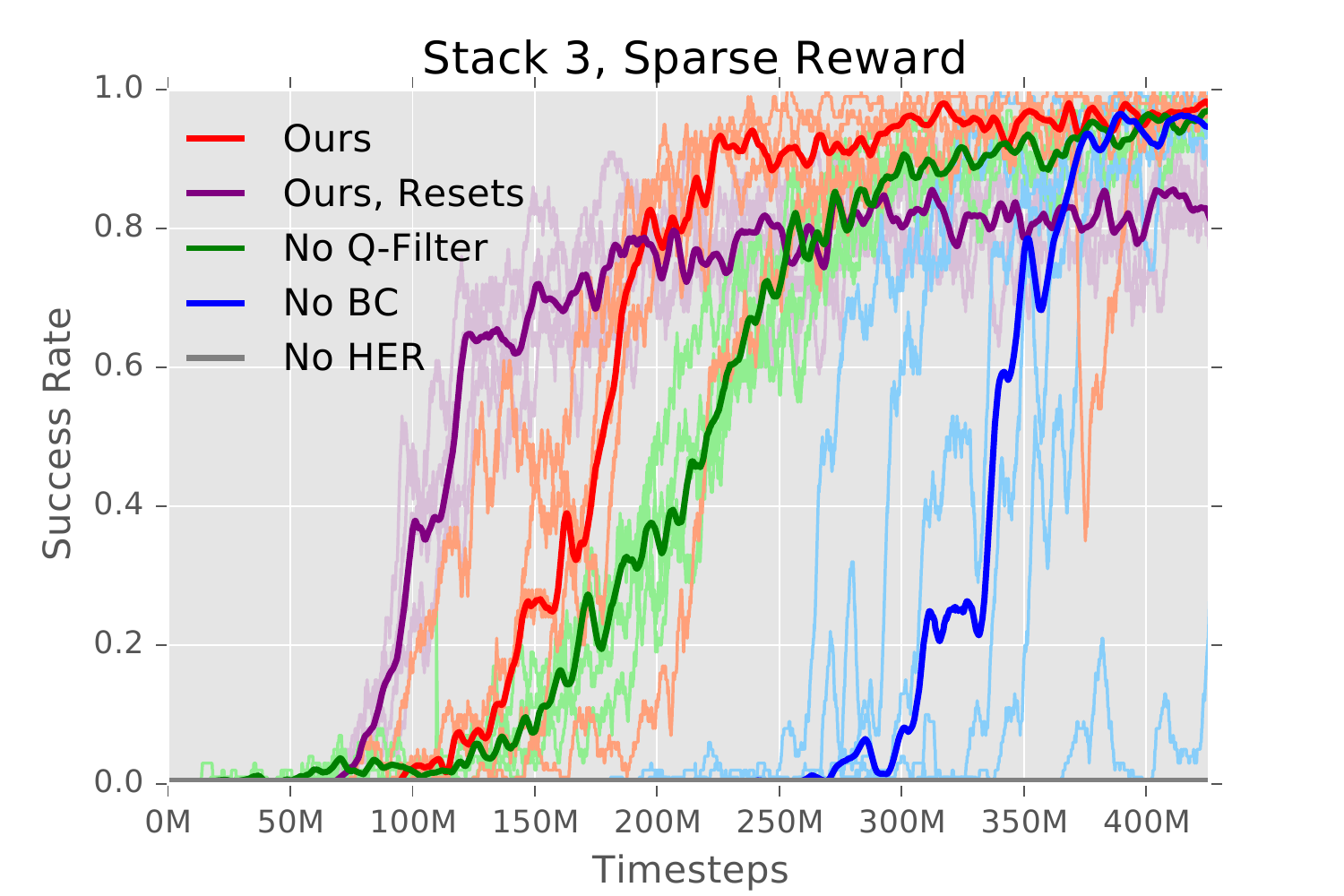}
    \caption{Ablation results on stacking 3 blocks with a fully sparse reward. We run each method 5 times with random seeds. The bold line shows the median of the 5 runs while each training run is plotted in a lighter color. Note ``No HER'' is always at 0\% success rate. Our method without resets learns faster than the ablations. Our method with resets initially learns faster but converges to a worse success rate. }%
    \label{fig:ablation3stack}%
\end{figure}

\begin{figure*}[t]%
    \vspace{6pt}
    \centering
    \begin{subfigure}{0.3\linewidth}
        \includegraphics[width=1.0\linewidth]{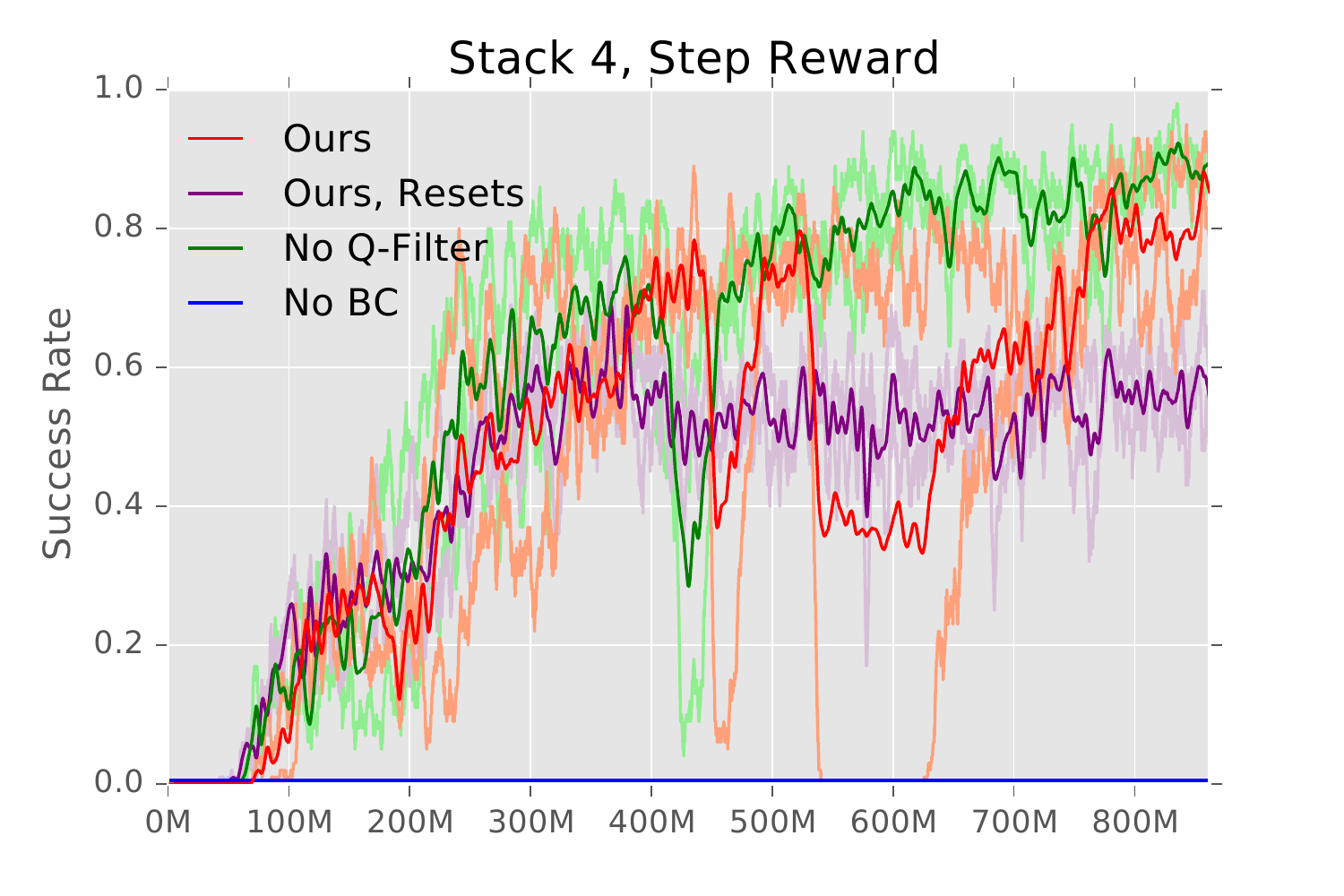}
        \includegraphics[width=1.0\linewidth]{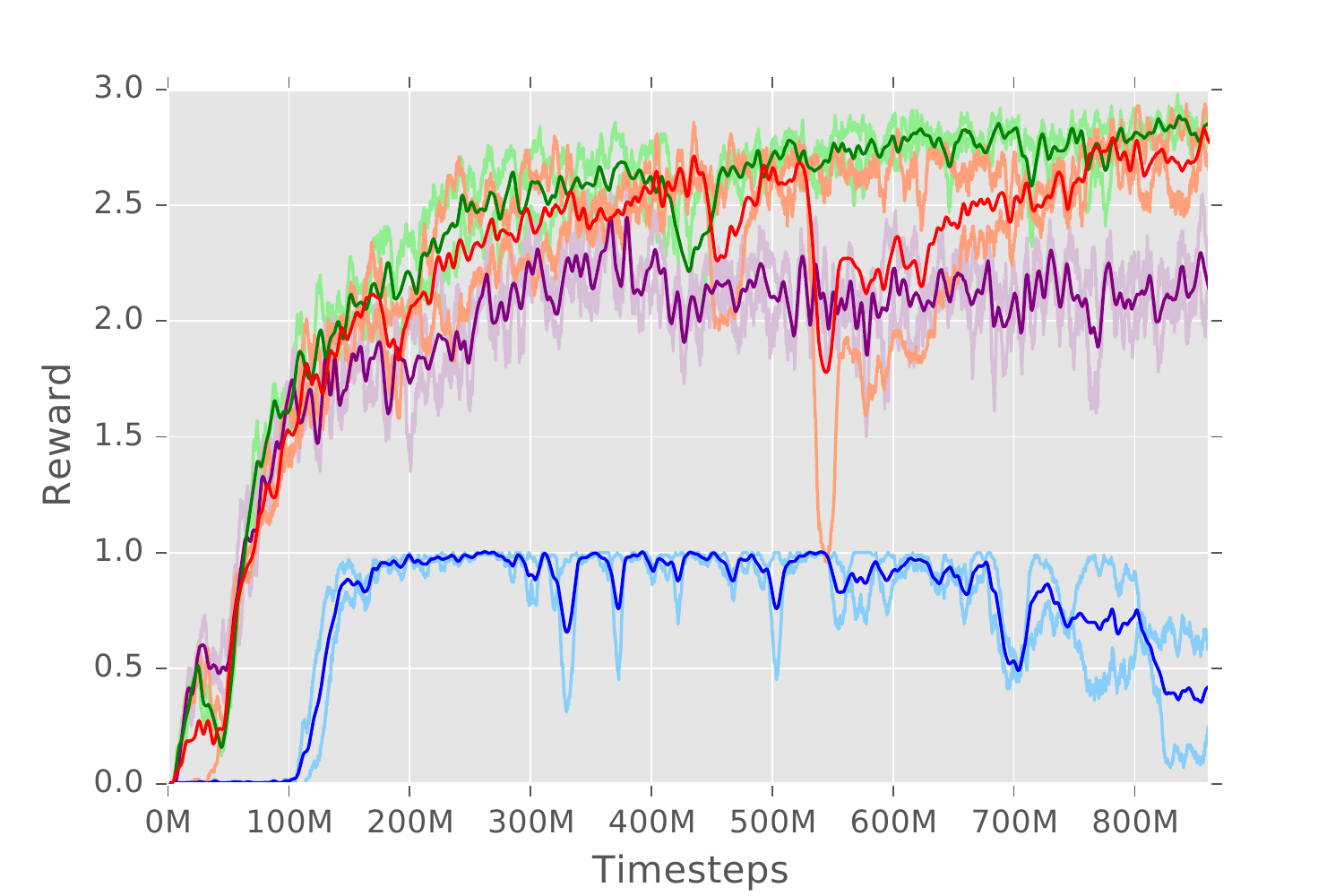}
    \end{subfigure}
    \begin{subfigure}{0.3\linewidth}
        \includegraphics[width=1.0\linewidth]{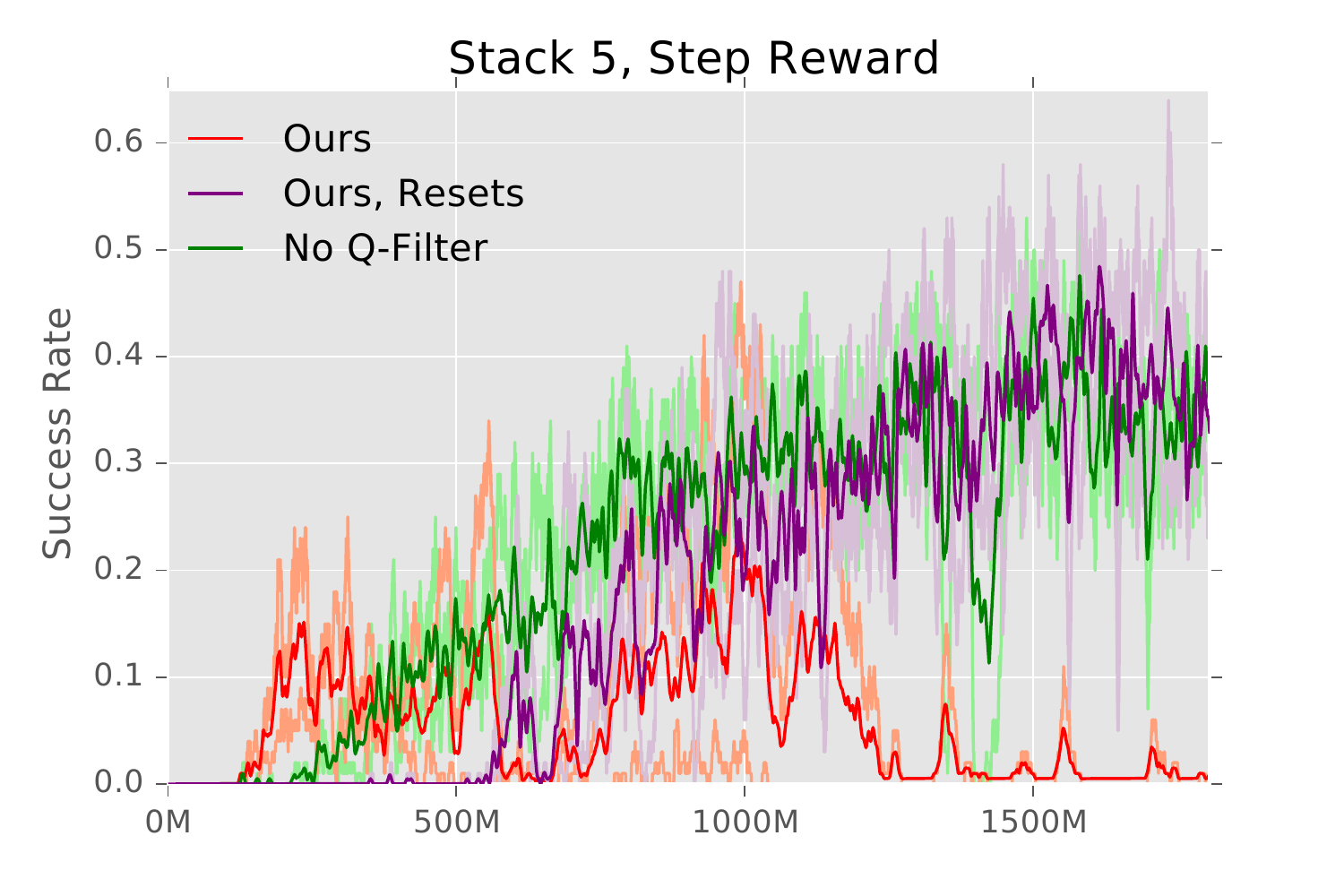}
        \includegraphics[width=1.0\linewidth]{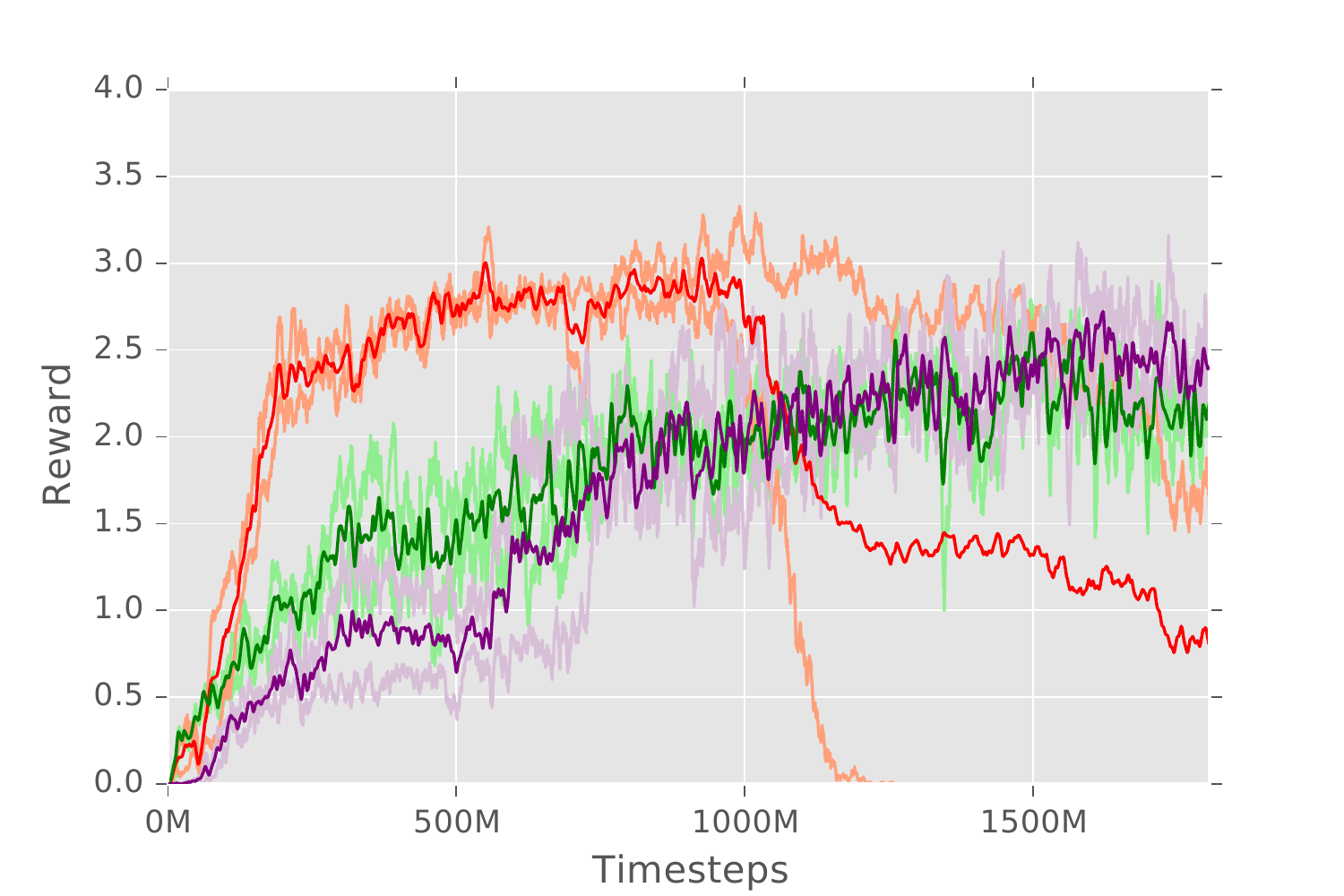}
    \end{subfigure}
    \begin{subfigure}{0.3\linewidth}
        \includegraphics[width=1.0\linewidth]{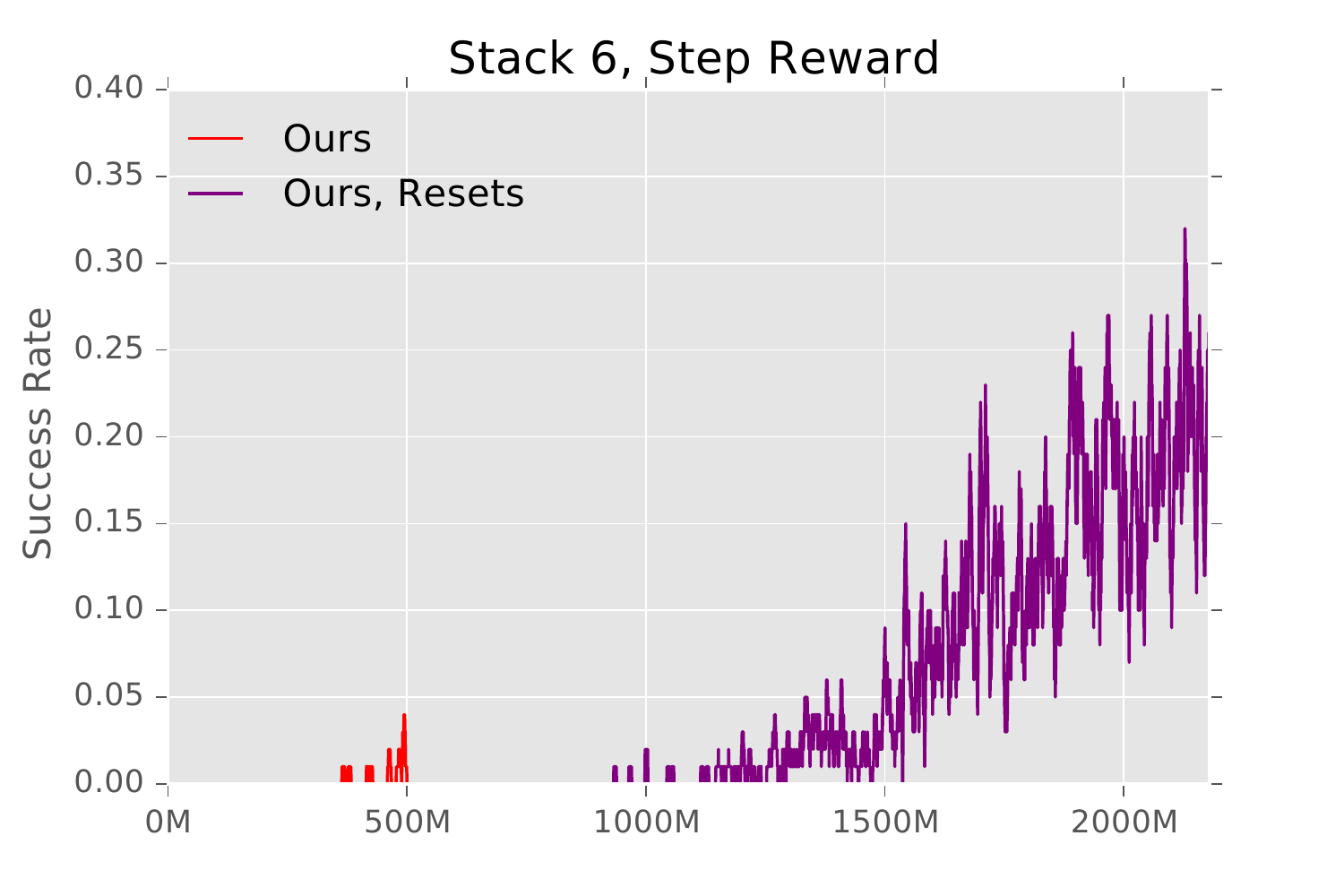}
        \includegraphics[width=1.0\linewidth]{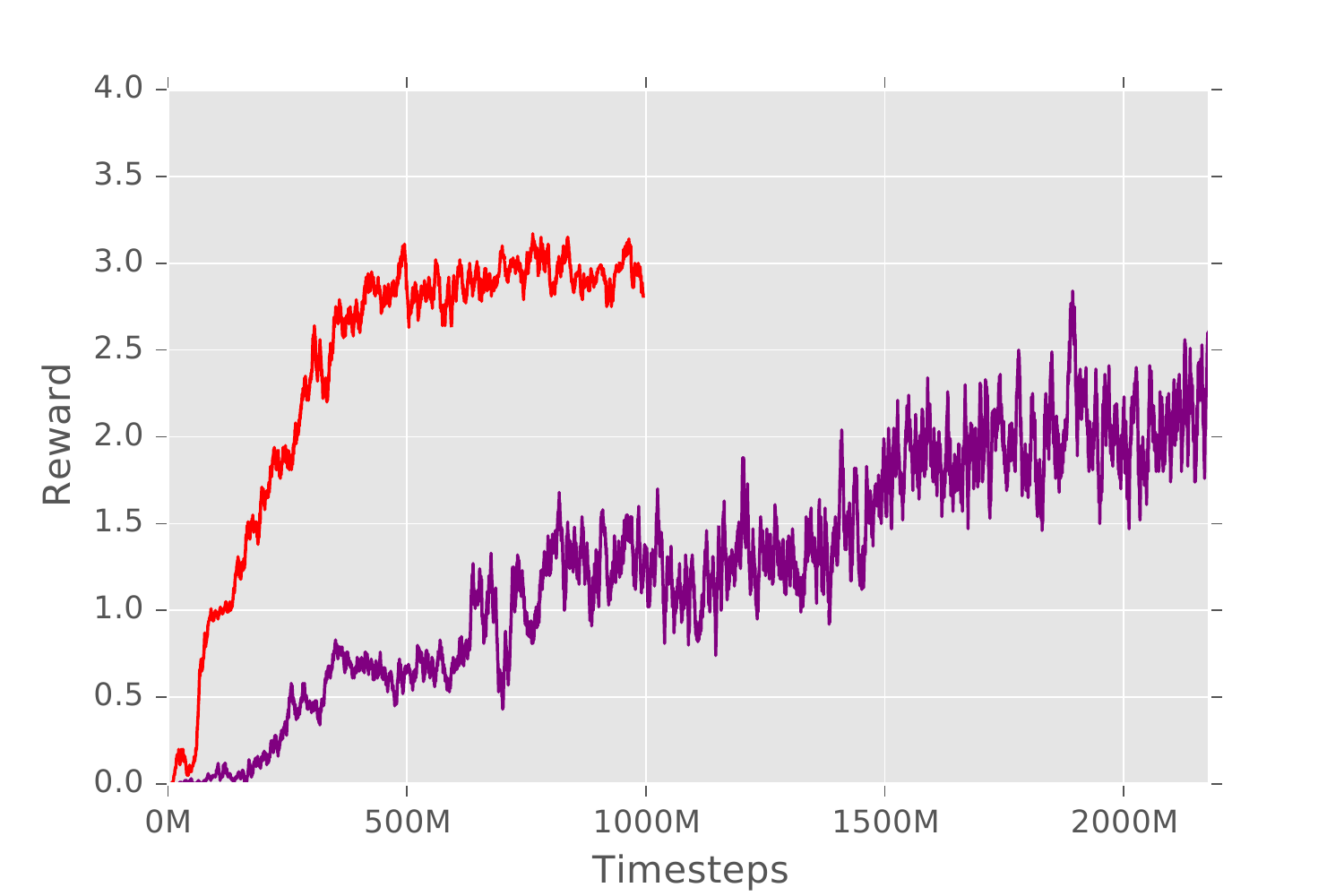}
    \end{subfigure}
    
    \caption{Ablation results on longer horizon tasks with a step reward. The upper row shows the success rate while the lower row shows the  average reward at the final step of each episode obtained by different algorithms. For stacking 4 and 5 blocks, we use 2 random seeds per method. The median of the runs is shown in bold and each training run is plotted in a lighter color. Note that for stacking 4 blocks, the ``No BC'' method is always at 0\% success rate. As the number of blocks increases, resets from demonstrations becomes more important to learn the task. }%
    \label{fig:step1}%
\end{figure*}

\section{Ablation Experiments}
\label{sec:ablations}

In this section we perform a series of ablation experiments to measure the importance of various components of our method. We evaluate our method on stacking 3 to 6 blocks.

We perform the following ablations on the best performing of our models on each task:

\noindent \textbf{No BC Loss:} This method does not apply the behavior cloning gradient during training. It still has access to demonstrations through the demonstration replay buffer.

\noindent \textbf{No Q-Filter:} This method uses standard behavioral cloning loss instead of the loss from equation Eq.~\ref{eq:filter}, which means that the actor tries to mimic the demonstrator's behaviour
regardless of the critic.

\noindent \textbf{No HER:} Hindsight Experience Replay is not used.

\subsection{Behavior Cloning Loss}

Without the behavior cloning loss, the method is significantly worse in every task we try. Fig. \ref{fig:ablation3stack} shows the training curve for learning to stack 3 blocks with a fully sparse reward. Without the behavior cloning loss, the system is about 2x slower to learn. On longer horizon tasks, we do not achieve any success without this loss.

To see why, consider the training curves for stacking 4 blocks shown in Fig. \ref{fig:step1}. The ``No BC'' policy learns to stack only one additional block. Without the behavior cloning loss, the agent only has access to the demonstrations through the demonstration replay buffer. This allows it to view high-reward states and incentivizes the agent to stack more blocks, but there is a stronger disincentive: stacking the tower higher is risky and could result in lower reward if the agent knocks over a block that is already correctly placed. Because of this risk, which is fundamentally just another instance of the agent finding a local optimum in a shaped reward, the agent learns the safer behavior of pausing after achieving a certain reward. Explicitly weighting behavior cloning steps into gradient updates forces the policy to continue the task.

\subsection{Q-Filter} 

The Q-Filter is effective in accelerating learning and achieving optimal performance. Fig. \ref{fig:ablation3stack} shows that the method without filtering is slower to learn. One issue with the behavior cloning loss is that if the demonstrations are suboptimal, the learned policy will also be suboptimal. Filtering by Q-value gives a natural way to anneal the effect of the demonstrations as it automatically disables the BC loss when a better action is found. However, it gives mixed results on the longer horizon tasks. One explanation is that in the step reward case, learning relies less on the demonstrations because the reward signal is stronger. Therefore, the training is less affected by suboptimal demonstrations.

\subsection{Resets From Demonstrations}

We find that initializing rollouts from within demonstration states greatly helps to learn to stack 5 and 6 blocks but hurts training with fewer blocks, as shown in Fig. \ref{fig:step1}. Note that even where resets from demonstration states helps the final success rate, learning takes off faster when this technique is not used. However, since stacking the tower higher is risky, the agent learns the safer behavior of stopping after achieving a certain reward. Resetting from demonstration states alleviates this problem because the agent regularly experiences higher rewards.

This method changes the sampled state distribution, biasing it towards later states. It also inflates the Q values unrealistically. Therefore, on tasks where the RL algorithm does not get stuck in solving a subset of the full problem, it could hurt performance.

\section{Discussion and Future Work}

We present a system to utilize demonstrations along with reinforcement learning to solve complicated multi-step tasks. We believe this can accelerate learning of many tasks, especially those with sparse rewards or other difficulties in exploration. Our method is very general, and can be applied on any continuous control task where a success criterion can be specified and demonstrations obtained.

An exciting future direction is to train policies directly on a physical robot. Fig. \ref{fig:baseline1} shows that learning the pick-and-place task takes about 1 million timesteps, which is about 6 hours of real world interaction time. This can realistically be trained on a physical robot, short-cutting the simulation-reality gap entirely. Many automation tasks found in factories and warehouses are similar to pick-and-place but without the variation in initial and goal states, so the samples required could be much lower. With our method, no expert needs to be in the loop to train these systems: demonstrations can be collected by users without knowledge about machine learning or robotics and rewards could be directly obtained from human feedback.

A major limitation of this work is sample efficiency on solving harder tasks. While we could not solve these tasks with other learning methods, our method requires a large amount of experience which is impractical outside of simulation. To run these tasks on physical robots, the sample efficiency will have to improved considerably. We also require demonstrations which are not easy to collect for all tasks. If demonstrations are not available but the environment can be reset to arbitrary states, one way to learn goal-reaching but avoid using demonstrations is to reuse successful rollouts as in  \cite{florensa2017resets}.

Finally, our method of resets from demonstration states requires the ability to reset to arbitrary states. Although we can solve many long-horizon tasks without this ability, it is very effective for the hardest tasks. Resetting from demonstration rollouts resembles curriculum learning: we solve a hard task by first solving easier tasks. If the environment does not afford setting arbitrary states, then other curriculum methods will have to be used.

\section{Acknowledgements}

\noindent We thank Vikash Kumar and Aravind Rajeswaran for valuable discussions. We thank Sergey Levine, Chelsea Finn, and Carlos Florensa for feedback on initial versions of this paper. Finally, we thank OpenAI for providing a supportive research environment.

{\small
\bibliographystyle{IEEEtran}
\bibliography{rope}

\begin{thebibliography}{10}
\providecommand{\url}[1]{#1}
\csname url@samestyle\endcsname
\providecommand{\newblock}{\relax}
\providecommand{\bibinfo}[2]{#2}
\providecommand{\BIBentrySTDinterwordspacing}{\spaceskip=0pt\relax}
\providecommand{\BIBentryALTinterwordstretchfactor}{4}
\providecommand{\BIBentryALTinterwordspacing}{\spaceskip=\fontdimen2\font plus
\BIBentryALTinterwordstretchfactor\fontdimen3\font minus
  \fontdimen4\font\relax}
\providecommand{\BIBforeignlanguage}[2]{{%
\expandafter\ifx\csname l@#1\endcsname\relax
\typeout{** WARNING: IEEEtran.bst: No hyphenation pattern has been}%
\typeout{** loaded for the language `#1'. Using the pattern for}%
\typeout{** the default language instead.}%
\else
\language=\csname l@#1\endcsname
\fi
#2}}
\providecommand{\BIBdecl}{\relax}
\BIBdecl

\bibitem{andrychowicz2017her}
M.~Andrychowicz \emph{et~al.}, ``Hindsight experience replay,'' in
  \emph{Advances in neural information processing systems}, 2017.

\bibitem{vecerik17ddpgfd}
M.~Ve{\v{c}}er{\'{i}}k \emph{et~al.}, ``{Leveraging Demonstrations for Deep
  Reinforcement Learning on Robotics Problems with Sparse Rewards},''
  \emph{arXiv preprint arxiv:1707.08817}, 2017.

\bibitem{deisenroth2011blocks}
M.~P. Deisenroth, C.~E. Rasmussen, and D.~Fox, ``{Learning to Control a
  Low-Cost Manipulator using Data-Efficient Reinforcement Learning},''
  \emph{Robotics: Science and Systems}, vol. VII, pp. 57--64, 2011.

\bibitem{duan2017oneshotimitation}
Y.~Duan \emph{et~al.}, ``One-shot imitation learning,'' in \emph{NIPS}, 2017.

\bibitem{pomerleau1989alvinn}
D.~A. Pomerleau, ``{Alvinn: An autonomous land vehicle in a neural network},''
  \emph{NIPS}, pp. 305--313, 1989.

\bibitem{bojarski2016nvidia}
M.~Bojarski \emph{et~al.}, ``{End to End Learning for Self-Driving Cars},''
  \emph{arXiv preprint arXiv:1604.07316}, 2016.

\bibitem{giusti15trails}
A.~Giusti \emph{et~al.}, ``{A Machine Learning Approach to Visual Perception of
  Forest Trails for Mobile Robots},'' in \emph{IEEE Robotics and Automation
  Letters.}, 2015, pp. 2377--3766.

\bibitem{nakanishi2004bipedlfd}
J.~Nakanishi \emph{et~al.}, ``{Learning from demonstration and adaptation of
  biped locomotion},'' in \emph{Robotics and Autonomous Systems}, vol.~47, no.
  2-3, 2004, pp. 79--91.

\bibitem{kalakrishnan09terraintemplates}
M.~Kalakrishnan \emph{et~al.}, ``{Learning Locomotion over Rough Terrain using
  Terrain Templates},'' in \emph{{The 2009 IEEE/RSJ International Conference on
  Intelligent Robots and Systems}}, 2009.

\bibitem{ross2011dagger}
S.~Ross, G.~J. Gordon, and J.~A. Bagnell, ``{A Reduction of Imitation Learning
  and Structured Prediction to No-Regret Online Learning},'' in
  \emph{Proceedings of the 14th International Conference on Artificial
  Intelligence and Statistics (AISTATS)}, 2011.

\bibitem{ng2000irl}
A.~Ng and S.~Russell, ``{Algorithms for Inverse Reinforcement Learning},''
  \emph{International Conference on Machine Learning (ICML)}, 2000.

\bibitem{ziebart2008maxent}
B.~D. Ziebart \emph{et~al.}, ``{Maximum Entropy Inverse Reinforcement
  Learning.}'' in \emph{AAAI Conference on Artificial Intelligence}, 2008, pp.
  1433--1438.

\bibitem{abbeel2004apprenticeship}
P.~Abbeel and A.~Y. Ng, ``Apprenticeship learning via inverse reinforcement
  learning,'' in \emph{ICML}, 2004, p.~1.

\bibitem{finn16guidedcostlearning}
C.~Finn, S.~Levine, and P.~Abbeel, ``{Guided Cost Learning: Deep Inverse
  Optimal Control via Policy Optimization},'' in \emph{ICML}, 2016.

\bibitem{peters2010reps}
J.~Peters, K.~M{\"{u}}lling, and Y.~Alt{\"{u}}n, ``{Relative Entropy Policy
  Search},'' \emph{Artificial Intelligence}, pp. 1607--1612, 2010.

\bibitem{deisenroth2011pilco}
M.~P. Deisenroth and C.~E. Rasmussen, ``Pilco: A model-based and data-efficient
  approach to policy search,'' in \emph{ICML}, 2011, pp. 465--472.

\bibitem{mnih2015human}
V.~Mnih \emph{et~al.}, ``Human-level control through deep reinforcement
  learning,'' \emph{Nature}, vol. 518, no. 7540, pp. 529--533, 2015.

\bibitem{Silver2016}
D.~Silver \emph{et~al.}, ``{Mastering the game of Go with deep neural networks
  and tree search},'' \emph{Nature}, vol. 529, no. 7587, pp. 484--489, Jan
  2016.

\bibitem{LevineFDA15}
S.~Levine \emph{et~al.}, ``End-to-end training of deep visuomotor policies,''
  \emph{CoRR}, vol. abs/1504.00702, 2015.

\bibitem{pinto2015supersizing}
L.~Pinto and A.~Gupta, ``Supersizing self-supervision: Learning to grasp from
  50k tries and 700 robot hours,'' \emph{arXiv preprint arXiv:1509.06825},
  2015.

\bibitem{levine2016learning}
S.~Levine \emph{et~al.}, ``Learning hand-eye coordination for robotic grasping
  with deep learning and large-scale data collection,'' \emph{arXiv preprint
  arXiv:1603.02199}, 2016.

\bibitem{Gu2016b}
S.~Gu \emph{et~al.}, ``{Deep Reinforcement Learning for Robotic Manipulation
  with Asynchronous Off-Policy Updates},'' \emph{arXiv preprint
  arXiv:1610.00633}, 2016.

\bibitem{lillicrap2015continuous}
T.~P. Lillicrap \emph{et~al.}, ``Continuous control with deep reinforcement
  learning,'' \emph{arXiv preprint arXiv:1509.02971}, 2015.

\bibitem{mnih2016asynchronous}
V.~Mnih \emph{et~al.}, ``Asynchronous methods for deep reinforcement
  learning,'' in \emph{ICML}, 2016.

\bibitem{schulman2015trpo}
J.~Schulman \emph{et~al.}, ``Trust region policy optimization,'' in
  \emph{Proceedings of the twenty-first international conference on Machine
  learning}, 2015.

\bibitem{winograd72shrdlr}
T.~Winograd, \emph{Understanding Natural Language}.\hskip 1em plus 0.5em minus
  0.4em\relax Academic Press, 1972.

\bibitem{Kaelbling2011}
L.~P. Kaelbling and T.~Lozano-Perez, ``{Hierarchical task and motion planning
  in the now},'' \emph{IEEE International Conference on Robotics and
  Automation}, pp. 1470--1477, 2011.

\bibitem{Kavraki1996}
L.~Kavraki \emph{et~al.}, ``{Probabilistic roadmaps for path planning in
  high-dimensional configuration spaces},'' \emph{IEEE transactions on Robotics
  and Automation}, vol.~12, no.~4, pp. 566--580, 1996.

\bibitem{srivastava14tamp}
S.~Srivastava \emph{et~al.}, ``{Combined Task and Motion Planning Through an
  Extensible Planner-Independent Interface Layer},'' in \emph{International
  Conference on Robotics and Automation}, 2014.

\bibitem{popov17stacking}
I.~Popov \emph{et~al.}, ``{Data-efficient Deep Reinforcement Learning for
  Dexterous Manipulation},'' \emph{arXiv preprint arXiv:1704.03073}, 2017.

\bibitem{schaal97lfd}
S.~Schaal, ``{Robot learning from demonstration},'' \emph{Advances in Neural
  Information Processing Systems}, no.~9, pp. 1040--1046, 1997.

\bibitem{peters2008baseball}
J.~Peters and S.~Schaal, ``{Reinforcement learning of motor skills with policy
  gradients},'' \emph{Neural Networks}, vol.~21, no.~4, pp. 682--697, 2008.

\bibitem{kober2008mp}
J.~Kober and J.~Peter, ``{Policy search for motor primitives in robotics},'' in
  \emph{Advances in neural information processing systems}, 2008.

\bibitem{hester17dqfd}
T.~Hester \emph{et~al.}, ``{Learning from Demonstrations for Real World
  Reinforcement Learning},'' \emph{arXiv preprint arxiv:1704.03732}, 2017.

\bibitem{kim2013apid}
B.~Kim \emph{et~al.}, ``{Learning from Limited Demonstrations},'' \emph{Neural
  Information Processing Systems.}, 2013.

\bibitem{schaul2015uva}
T.~Schaul \emph{et~al.}, ``{Universal Value Function Approximators},''
  \emph{Proceedings of The 32nd International Conference on Machine Learning},
  pp. 1312--1320, 2015.

\bibitem{todorov12mujoco}
E.~Todorov, T.~Erez, and Y.~Tassa, ``{MuJoCo: A physics engine for model-based
  control},'' in \emph{The IEEE/RSJ International Conference on Intelligent
  Robots and Systems}, 2012.

\bibitem{kingma2014adam}
D.~Kingma and J.~Ba, ``Adam: A method for stochastic optimization,''
  \emph{International Conference on Learning Representations (ICLR)}, 2015.

\bibitem{bahdanau14attention}
D.~Bahdanau, K.~Cho, and Y.~Bengio, ``{Neural Machine Translation by Jointly
  Learning to Align and Translate},'' in \emph{ICLR}, 2015.

\bibitem{florensa2017resets}
C.~Florensa \emph{et~al.}, ``{Reverse Curriculum Generation for Reinforcement
  Learning},'' in \emph{Conference on robot learning}, 2017.

\end{thebibliography}
}

\end{document}